\documentclass[11pt,a4paper]{article}
\usepackage[hyperref]{conll2020}
\usepackage{times}
\usepackage{latexsym}

\usepackage{microtype}
\usepackage{linguex}
\usepackage{hyperref}
\usepackage{caption,subcaption}
\usepackage{epsfig}

\aclfinalcopy %
\def\aclpaperid{144} %

\newif\ifparens\parensfalse

\newcommand\pref[1]{{\parenstrue\ref{#1}}}

\title{Discourse structure interacts with reference but not syntax in neural language models}

\author{Forrest Davis \and Marten van Schijndel \\
        Department of Linguistics\\
        Cornell University\\
        \texttt{\{fd252|mv443\}@cornell.edu}}

\date{}

\begin{document}
\maketitle
\begin{abstract}
 Language models (LMs) trained on large quantities of text have been claimed to acquire abstract linguistic representations. Our work tests the robustness of these abstractions by focusing on the ability of LMs to learn interactions between different linguistic representations. In particular, we utilized stimuli from psycholinguistic studies showing that humans can condition reference (i.e.\ coreference resolution) and syntactic processing on the same discourse structure (implicit causality). We compared both transformer and long short-term memory LMs to find that, contrary to humans, implicit causality only influences LM behavior for reference, not syntax, despite model representations that encode the necessary discourse information. Our results further suggest that LM behavior can contradict not only learned representations of discourse but also syntactic agreement, pointing to shortcomings of standard language modeling.
\end{abstract}

\section{Introduction} \label{sec:intro}

Neural network language models (LMs), pretrained on vast amounts of raw text, have become 
the dominant input to downstream tasks \cite{petersetal18,radfordetal18,devlinetal19}. Commonly, these tasks involve aspects of language 
comprehension (or understanding). One explicit example is coreference resolution, wherein anaphora (e.g., pronouns) are linked to antecedents (e.g., nouns) requiring knowledge of syntax, semantics, 
and world-knowledge to match human-like comprehension. 

Recent work has suggested that LMs acquire abstract, often human-like, knowledge of syntax 
\cite[e.g.,][]{gulordavaetal18, futrelletal2018, huetal2020-systematic}. Additionally, knowledge
of grammatical and referential aspects linking a pronoun to its antecedent noun (reference) 
have been demonstrated for both 
transformer and long short-term memory architectures \cite{sorodoc-etal-2020-probing}. Humans are able 
to modulate both referential and syntactic comprehension 
given abstract linguistic knowledge (e.g., discourse structure). Contrary to humans, we find
that discourse structure (at least as it 
pertains to implicit causality) only influences LM behavior 
for reference, not syntax, despite model representations
that encode the necessary discourse information.

The particular discourse structure we examined is governed by
implicit causality (IC) verbs \cite{garvey1974implicit}. Such verbs influence pronoun comprehension:

\ex. \label{ic_exp}
    \a. Sally frightened Mary because she was so terrifying. \label{ic_exp_a}
    \b. Sally feared Mary because she was so terrifying. \label{ic_exp_b}

In \ref{ic_exp}, \textit{she} agrees in gender with both \textit{Sally} and \textit{Mary}, so 
both are possible antecedents. However, English speakers overwhelmingly 
interpret \textit{she} as referring to \textit{Sally} in \ref{ic_exp_a} and \textit{Mary} 
in \ref{ic_exp_b}, despite the semantic overlap between the verbs. Verbs that 
have a subject preference (e.g., \textit{frightened}) are called subject-biased IC verbs, and
verbs with a object preference (e.g., \textit{feared}) are called object-biased IC verbs. 

In addition to pronoun resolution, IC verbs also interact with relative clause (RC) attachment: 

\ex. \label{ic_rc}
    \a. \label{ic_rc_a} John babysits the children of the musician who...
        \a. \label{ic_rc_a_1} ...lives in La Jolla.
        \b. \label{ic_rc_a_2} ...are students at a private school.
        \z.
    \b. \label{ic_rc_b} John detests the children of the musician who...
        \a. \label{ic_rc_b_1} ...lives in La Jolla.
        \b. \label{ic_rc_b_2} ...are arrogant and rude. 
        \z.
    \z.
    \citep[from][]{rohdeetal2011}

In \ref{ic_rc}, \ref{ic_rc_a} and \ref{ic_rc_b} are sentence fragments with possible 
continuations modifying \textit{the musician} in \ref{ic_rc_a_1} and \ref{ic_rc_b_1} and 
continuations modifying \textit{the children} in \ref{ic_rc_a_2} and \ref{ic_rc_b_2}. We might expect 
human continuation preferences to be the same in \ref{ic_rc_a} and \ref{ic_rc_b}. However, the use 
of an object-biased IC verb (\textit{detests}) in \ref{ic_rc_b} increases the proportion of continuations given by human participants 
that refer to the \textit{children} (i.e.\ \ref{ic_rc_b_2} vs.\ \ref{ic_rc_b_1}). Without 
an object-biased IC verb the majority of continuations refer to the more recent noun 
(i.e.\ \textit{musician}).

Effects 
of IC have received renewed interest in the field of psycholinguistics in recent years \cite[e.g.,][]{kehler2008coherence, ferstl2011implicit, hartshorne2013verb, hartshorne2014, williams_IC_2020}. Current accounts of IC claim that the phenomenon is inherently a linguistic process, which 
does not rely on additional pragmatic inferences by comprehenders \cite[e.g.,][]{rohdeetal2011, hartshorne2013verb}. Thus, IC is argued to be contained within the linguistic signal, analogous to 
evidence of
syntactic agreement and verb argument structure within corpora. We 
hypothesize that if these claims are correct, then current LMs will be able to 
condition reference and syntactic attachment by 
IC verbs with just language data (i.e.\ without grounding). 

We tested this hypothesis using unidirectional transformer and long short-term memory network \citep[LSTM;][]{hochreiterschmidhuber97} language models. We find that LSTM 
LMs fail to acquire a subject/object-biased IC distinction that influences reference or RC attachment.  
In contrast, transformers learned a representational 
distinction between subject-biased and object-biased IC verbs that interacts 
with both reference and RC attachment, 
but the distinction only influenced model output for reference. The apparent failure of model 
syntactic behavior to exhibit an IC 
contrast that is present in model
representations raises questions 
about the broader capacity of LMs to display 
human-like linguistic knowledge.

\section{Related Work}\label{related}
The ability of LMs to encode referential knowledge has largely 
been explored in the domain of coreference resolution. Prior 
work has suggested that LMs can learn coreference resolution to 
some extent \cite[e.g.,][]{petersetal18, sorodoc-etal-2020-probing}.
In the present study, we focus
on within-sentence resolution 
rather than the ability of LMs to track entities over larger 
spans of text \cite[cf.][]{sorodoc-etal-2020-probing}. Previous 
work at this granularity of coreference resolution
has shown LSTM LMs strongly 
favor reference to male entities \cite{jumelet-etal-2019-analysing}, 
for which the present study finds additional support. Rather 
than utilizing a more limited modeling objective such as coreference resolution \cite[cf.][]{cheng-erk-2020-attending}, 
we followed \citet{sorodoc-etal-2020-probing} in focusing on the 
representation of referential knowledge by models trained with 
a general language modeling objective. 

With regards to linguistic representations, a growing body of literature 
suggests that LSTM LMs are able to acquire syntactic 
knowledge. In particular, subject-verb agreement has been explored extensively
\cite[e.g.,][]{linzenetal16, bernardyetal2017, Enguehard17}
with results at human level performance in some cases \cite{gulordavaetal18}. Additionally, work has shown human-like
behavior when processing reflexive pronouns, 
negative polarity items \cite{futrelletal2018}, center embedding, and syntactic islands \cite{wilcox2019block, wilcox2019supression}. This
literature generally suggests that LMs encode some type of abstract 
syntactic representation \cite[e.g.,][]{prasadetal2019}. Additionally, recent work has shown 
LMs learn linguistic representations beyond syntax, such 
as pragmatics and discourse structure 
\cite{jeretic-etal-2020-natural,schuster-etal-2020-harnessing, davis-van-schijndel-2020b}.

The robustness of these abstract linguistic representations, however, 
have been questioned in recent work, suggesting that learned 
abstractions are weaker than standardly assumed \cite[e.g.,][]{trasketal2018, vanSchijndeletal2019, kodner-gupta-2020-overestimation, davis-van-schijndel-2020-recurrent}. The
present study builds on these recent developments by demonstrating 
the inability of LMs to utilize discourse structure in syntactic 
processing. 

\section{Language Models}
We trained 25 LSTM LMs on the Wikitext-103 
corpus \cite{merityetal16} with a
vocabulary constrained to the most frequent 50K words.\footnote{The models had two LSTM layers with 400 hidden units each, 400-dimensional word embeddings, a dropout rate of 0.2 and batchsize 20, and were trained for 40 epochs (with early stopping) using PyTorch. The mean perplexity for the models
on the validation data was 40.6 with a standard deviation of 2.05. The LSTMs and code for the experiments in this paper can be found at \url{https://github.com/forrestdavis/ImplicitCausality}.} We used two pretrained unidirectional
transformer LMs: TransformerXL \cite{dai-etal-2019-transformer} and GPT-2 XL \cite{radfordetal19}.\footnote{We used HuggingFace's implementation of these 
models \cite{Wolf2019HuggingFacesTS}.} 

TransformerXL was trained on Wikitext-103, like our 
LSTM LMs, but has more parameters and a larger vocabulary.  
GPT-2 XL differs from the other models in lacking recurrence (instead utilizing non-recurrent 
self-attention) and in amount and diversity of training data
(1 billion words compared to the 103 million
in Wikitext-103). As such, we caution against extracting 
explicit, mechanistic claims from the present study
concerning the relationship between learned linguistic knowledge and model configurations and 
training data. Instead, our work points 
to apparent differences between transformers and LSTMs with regard to use and acquisition of 
discourse structure, leaving explanatory principles to further work. 

\section{Interactions with Reference} \label{sec:ref}

The results of \citet{sorodoc-etal-2020-probing} suggested that 
referential contrasts based in grammatical features (e.g., gender) 
would be easier for models to discern then those purely focused 
on referential selection (e.g., antecedents with the same gender but differing
in preference). To evaluate this claim, we analyzed the degree 
to which IC verb type (i.e.\ subject vs.\ object biased) influenced i) model 
pronoun preferences when the possible referents differed in gender (e.g., \textit{Sally
feared Bob because...}), and ii) similarity of model representations between the pronoun 
and possible referents when they share the same gender (e.g., \textit{Fred feared Bob because he...}). Our prediction
was that IC would have a weaker 
influence in (ii) than (i). 

\subsection{Referential Stimuli}

Our data consisted of the stimuli from a human experiment conducted in \citet{ferstl2011implicit}, which asked participants to give continuations 
to sentence fragments of the following form: 

\ex. \label{ferstl}
Kate accused Bill because ... 

Continuations were coded across 305 verbs for whether participants
referenced the subject (i.e.\ \textit{she}) or the object (i.e.\ \textit{he}).\footnote{An additional category, other, was included for ambiguous (i.e.\ \textit{they hate each other}) or non-referential continuations (i.e.\ \textit{it was a rough day}).} The results of this coding were then 
converted into a bias score for each verb, ranging from 100 for verbs whose valid continuations 
uniquely refer to the subject (i.e.\ subject-biased) 
to -100 for verbs whose valid continuations uniquely refer 
to the object (i.e.\ object-biased). In the present study, we took 246 of these verbs\footnote{59 
verbs were outside of our LSTM LM vocabulary, so they were excluded.} and generated
stimuli as in \ref{ferstl} using 14 pairs of stereotypical male and female nouns (e.g., \textit{man} vs. \textit{woman}, \textit{king} vs. \textit{queen}), rather than rely on proper names as was done in 
\citet{ferstl2011implicit}.\footnote{See Appendix \ref{pairs} for all the pairs.} We created two categories of stimuli, those with differing gender\footnote{We balanced our stimuli by gender, so we had the same number of female 
subjects as male subjects and vice versa.} and those with the same gender resulting in 6888 sentences per category.  

\subsection{Measures} \label{sec:measures}

We evaluated our models independently for external behavior (e.g., predicted next-words) and internal 
representations (e.g., hidden states). Ideally, model 
behavior should 
condition on an abstracted representation that 
distinguishes subject-biased IC verbs from 
object-biased IC verbs. Similarly, a representational 
distinction between subject and object-biased IC verbs
should have some influence on model behavior. We find that this is not the case for
the LMs under investigation; a representational 
distinction between subject vs.\ object-biased IC verbs 
does not condition model behavioral differences.

To evaluate behavior, we appended a 
pronoun to \ref{ferstl} and calculated information-theoretic surprisal \cite{shannon48, hale2001probabilistic, levy2008expectation}. Surprisal is defined as the inverse log probability
assigned to each word ($w_i$) in a sentence given the preceding context: %
\begin{equation}
\textrm{surprisal}(w_i) = -\textrm{log}\,p(w_i|w_1 ... w_{i-1})%
\end{equation}

The probability of a word was calculated by applying the softmax function to a LM's output layer. Surprisal 
has been correlated with human processing difficulty \cite{smith2013effect,franketal15} allowing us to compare model behavior 
to human behavior. We predicted that IC verbs would influence surprisal, 
with subject-biased ICs lowering the surprisal of pronouns agreeing with the subject, and 
object-biased ICs lowering the surprisal of the prounouns agreeing with the object. This methodology follows subject-verb agreement experiments,
where verbs that agree in number with the subject
are less surprising than those that do not \cite[e.g., \textit{Cats are} vs.\ \textit{*Cats is}; ][]{linzenetal16, mueller-etal-2020-cross} 

To evaluate model representations, we followed work in 
representational similarity analysis
and used Pearson's r\footnote{Specifically, \href{https://numpy.org/doc/stable/reference/generated/numpy.corrcoef.html}{corrcoef} from numpy.} to measure
the similarity between model representations \cite[see][]{kriegeskorte2008representational, chrupala-alishahi-2019-correlating}.\footnote{There exist a number of other measures of representational 
similarity \cite[e.g.,][]{morcos2018insights}. In the present study, our use of experimental materials from psycholinguistic studies 
resulted in far fewer data than is needed 
for these methods, where one wants 
much more data than the dimensions of the representations. This is particularly stark for the syntactic stimuli where 
the embedding size for GPT-2 XL is roughly 13 times larger than the number of stimuli. These techniques may ultimately provide stronger evidence for representations of implicit causality in these language models, particularly for the LSTM LMs where
no representational trace of implicit causality was found. It is worth noting that the LSTM behavior does not show an influence of implicity causality, so if we were to find such a representation with a 
better measure of similarity it would further the disconnect between model representations and behavior we found for the transformers. We hope to 
explore this in future work.}
 
We build on work that has looked at the 
propagation of information across 
time within a layer \cite[as in][]{ giulianelli-etal-2018-hood, jumelet-etal-2019-analysing}. In such 
work, model behavior 
for subject-verb agreement and 
coreference is linked to model representations, and in particular, 
stronger model representations of previous
time steps that relate to the model's current prediction.

In the present study, we focused on the similarity 
between the 
pronoun and possible antecedents 
when they shared the same 
gender:

\ex. \label{same_gender}
  \a. \label{same_gender_a} The mother amused the girl because she ...
  \b. \label{same_gender_b} The mother applauded the girl because she ...

\begin{figure}[t] 
\includegraphics[width=\linewidth]{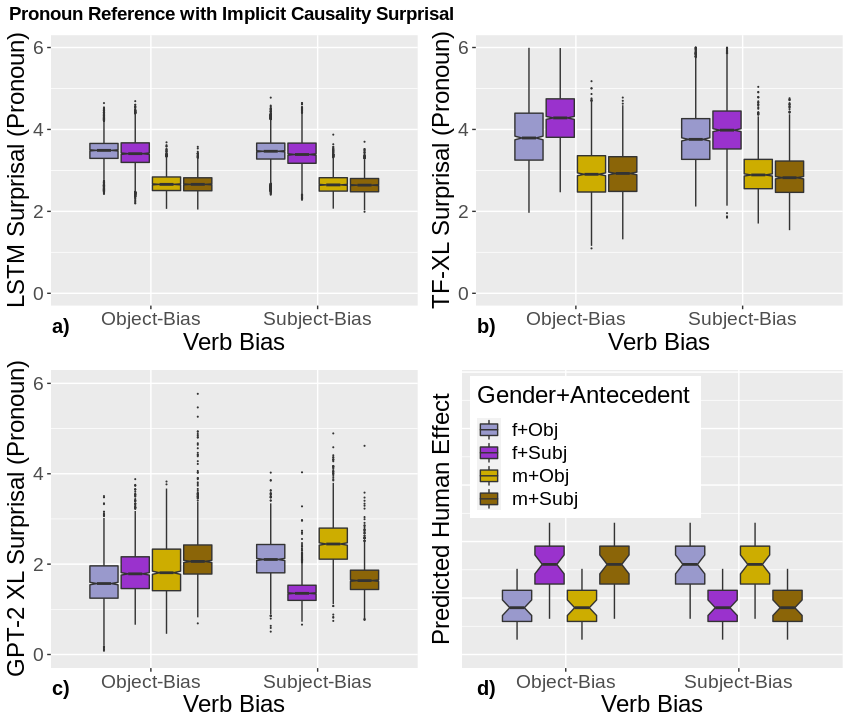}
\caption{Model surprisal (in \textbf{a)} LSTM LMs, \textbf{b)} TransformerXL, and \textbf{c)} GPT-2 XL) at the pronoun and \textbf{d)} the predicted 
qualitative human-like pattern; stimuli
from \citet{ferstl2011implicit} (e.g., \textit{the man accused 
the woman because she}). Broken into antecedent (subject vs.\ object) and gender of pronoun (male vs.\ female). Lower surprisal 
corresponds to greater model preference. }
\label{fig:pronoun_surp}
\end{figure}
    
Specifically, for \ref{same_gender}
we computed the layer-wise similarity between the hidden representation of \textit{she} with 
\textit{mother} and \textit{girl}.\footnote{Given the BPE tokenizer for GPT-2 XL, if a 
noun was broken into components, we used the hidden representation of the final 
component.} The bias score found in \citet{ferstl2011implicit} for \textit{amused} in \ref{same_gender_a}
was 67 (i.e.\ the verb is subject-biased) and for \textit{applauded} in \ref{same_gender_b} it was -84 (i.e.\ the verb is object-biased). Thus, we predicted that a layer that encodes a
human-like IC distinction should have greater similarity between
\textit{she} and \textit{mother} in the case of \textit{amused} than between 
\textit{she} and \textit{girl}, and vice versa
for \textit{applauded}.

\subsection{Influence of IC on Referential Behavior}

We calculated the surprisal for our LMs at the pronoun in our experimental 
stimuli, 
with the prediction that IC bias would modulate surprisal. Results 
for each LM type (LSTMs, TransformerXL, GPT-2 XL) are given in Figure \ref{fig:pronoun_surp}. 
Statistical analyses\footnote{We used lmer \cite[version 1.1.23;][]{bates2015lme4} and lmerTest \cite[version 3.1.2;][]{kuznetsova2017lmertest} in R.} were conducted via linear-mixed effects models.\footnote{We fit a model to predict
surprisal at the pronoun with a three way interaction between IC bias, position of gender matching 
noun, and gender of pronoun and a random intercept for item. We ran a model with the 
continuous bias score from \citet{ferstl2011implicit} and another with a categorical 
bias effect derived from the bias score in \citet{ferstl2011implicit}, with positive 
bias scores corresponding to a subject-biased verb and negative bias scores corresponding 
to a object-biased verb. These models had comparable results and 
are reported in the supplemental materials.} Post-hoc
t-tests were conducted to assess effects.\footnote{The threshold for statistical significance was p = 0.005. Full output from the statistical models are given in the supplemental materials, and all R code to recreate the tests and figures is on Github.}

 \begin{figure*}[t!]
\includegraphics[width=\textwidth]{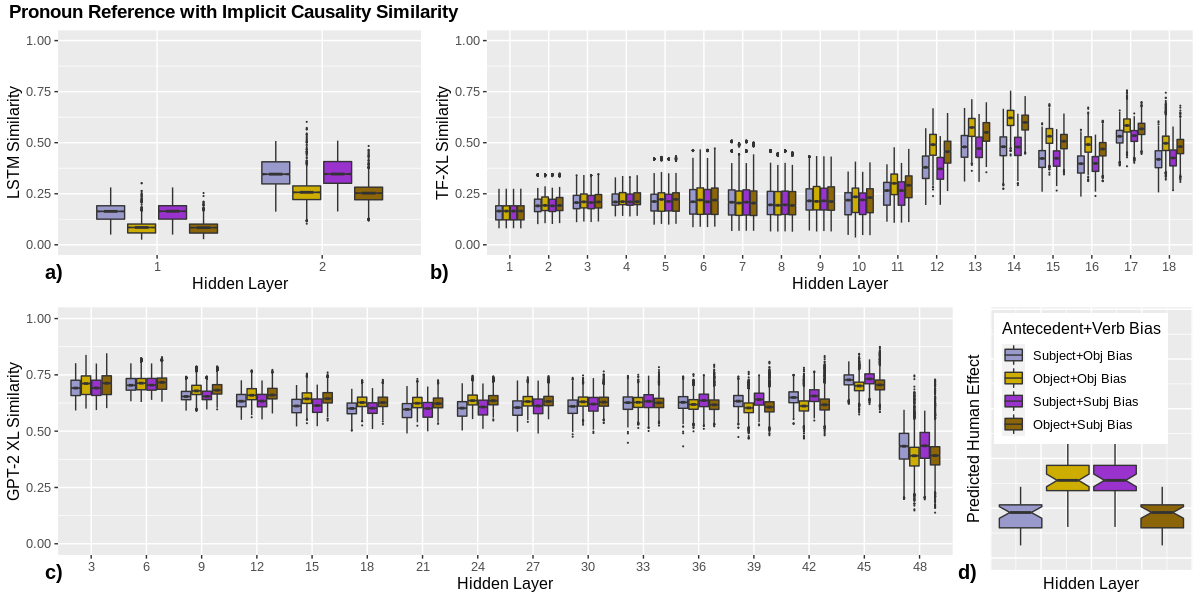}
\caption{Layer-wise representational similarity (in \textbf{a)} LSTM LMs, \textbf{b)} TransformerXL, and \textbf{c)} GPT-2 XL) between 
pronoun and subject/object and \textbf{d)} the predicted
qualitative human-like pattern); stimuli from \citet{ferstl2011implicit} (e.g., \textit{the man accused 
the boy because he}).
Broken into antecedent (subject vs.\ object) 
and IC bias type (subject-bias vs.\ object-bias). We include every third layer for GPT-2 XL (48 layers total). Greater similarity corresponds 
to greater relationship between pronoun and antecedent.}
\label{fig:pronoun_sim}
\end{figure*}

As is visually apparent in Figure \ref{fig:pronoun_surp}, all three models showed 
some gender bias (male for 
TransformerXL and LSTM LMs and female for GPT-2 XL), in line with existing findings of
gender preferences in LSTMs \cite[see][]{jumelet-etal-2019-analysing}. 

The effect of IC bias was mixed across the LMs. For the LSTMs, the influence 
of IC was marginal ($p = 0.02$) being driven by an
extremely small difference (0.02 bits) in surprisal
centered on male pronouns agreeing 
in gender with the subject. There was 
neither a significant effect for object pronouns or for female pronouns referring to subjects. 
We concluded that the LSTM IC effect was spurious and that
LSTM LMs acquired no 
IC-conditioned expectation about reference. 

For TransformerXL, there was a slight lowering in surprisal for reference to male subjects 
with subject-biased verbs, %
and a larger lowering in surprisal 
for reference to female subjects after subject-biased verbs. 
That is to say, subject-biased IC verbs did lower the surprisal
of pronouns referring to subjects, 
as predicted. 
However, there was no influence of IC when 
pronouns referred to the object. This suggests that preferences 
for local agreement in TransformerXL 
are much stronger than the influence of IC-bias, which only 
appears with subject-biased verbs. 

The behavior of GPT-2 XL was in line with the human findings 
from \citet{ferstl2011implicit}. Subject-biased 
verbs lowered the surprisal of pronouns referring to the subject, and 
object-biased verbs lowered the surprisal of pronouns referring to 
the object, regardless of gender. %
This suggests that 
GPT-2 XL has acquired a robust IC representation that influences 
expectations for pronominal reference.

\begin{figure*}[t]
\centering
\includegraphics[width=\textwidth]{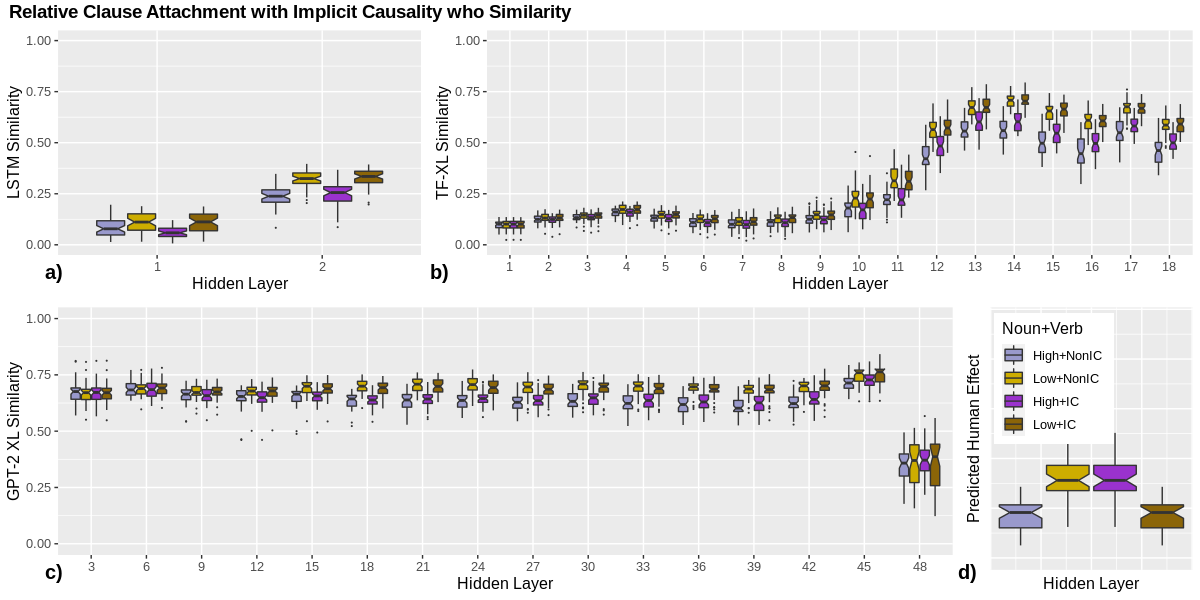}
\caption{Layer-wise representational similarity (in \textbf{a)} LSTM LMs, \textbf{b)} TransformerXL, and \textbf{c)} GPT-2 XL) between 
\textit{who} and the higher/lower noun, and \textbf{d)} the qualitative predicted human-like pattern); stimuli from \citet{rohdeetal2011} (e.g., \textit{the man admired the agent of the rockers who}).
Broken into attachment location (higher noun vs.\ lower noun) 
and verb type (object-biased IC verb vs.\ non-IC verb). 
We include every third layer for GPT-2 XL (48 layers total). Greater similarity corresponds 
to greater relationship between attachment location and \textit{who}.}
\label{fig:rc_sim}
\end{figure*}

\subsection{Influence of IC on Referential Representation}

We turn now to the ability of the models to distinguish 
the correct referent when both the subject and object 
have the same gender. Previous literature has suggested 
that this effect would be weaker than in the mismatching gender
case above \cite[see][]{sorodoc-etal-2020-probing}. We relied on 
a representational analysis (detailed in Section \ref{sec:measures}) 
to evaluate the preferences of the LMs. Results for each 
LM type are given in Figure \ref{fig:pronoun_sim}. 

Statistical significance was determined via linear-mixed 
effects models with post-hoc t-tests assessing the 
effects.\footnote{We fit models predicting 
similarity from a four way interaction of 
IC bias, noun comparison
(subject or object), model layer, and gender and a random 
intercept for item. Two IC bias effects were considered: the
gradient value given in \citet{ferstl2011implicit} and a categorical 
value where positive bias corresponds to a subject-biased verb 
and negative bias corresponds to a object-biased verb. Similar 
results were found with both effects with both 
models given in the supplemental materials.} As
predicted, IC bias had a weaker effect when 
choosing between competing nouns with the same
gender for reference (e.g., \textit{the woman admires the queen because she}). 

For LSTM LMs, IC bias did not influence 
model representations, at least as measured in the present study. For TransformerXL
there was a small difference in degree of similarity from 
layers 12 to 18. The pronoun 
was more similar to the object when the verb was object-biased.
In contrast, there was no significant effect for subject-biased verbs, despite 
the reverse effect in behavior when 
antecedents had mismatched gender (i.e.\ subject-bias, not object-bias, influenced pronoun surprisal in our 
behavioral analysis).

For GPT-2 XL we found a small, yet significant, 
difference in degree of similarity with the subject antecedent
starting in layer 15 and continuing through layer 47. That is, 
there was greater similarity between the pronoun and 
the subject when the verb was subject-biased. There was
no effect for pronouns referring to the object.
These results suggest that the influence of IC 
is only weakly present when both the subject and 
object are possible antecedents (i.e.\ they are 
the same gender). It therefore seems that models 
were only able to fully leverage an IC contrast to resolve 
reference when 
gender differences unambiguously 
distinguished between subject and object. 

\begin{figure}[t] 
\includegraphics[width=\linewidth]{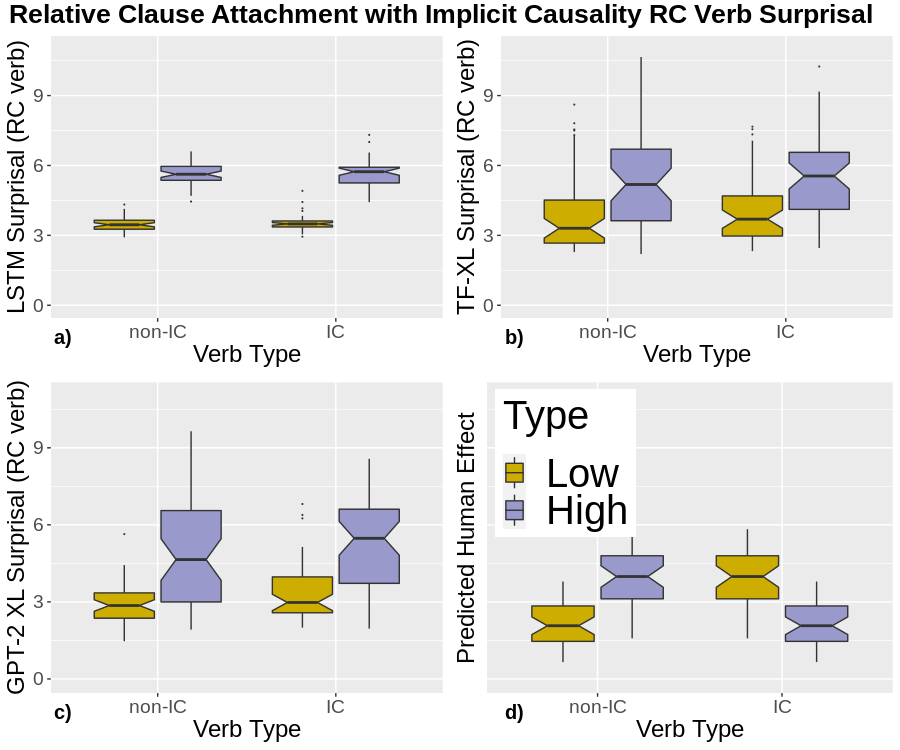}
\caption{Model surprisal (in \textbf{a)} LSTM LMs, \textbf{b)} TransformerXL, \textbf{c)} GPT-2 XL, 
and \textbf{d)} qualitative predicted human-like pattern) at the RC verb (\textit{was/were}); stimuli
from \citet{rohdeetal2011} (e.g., \textit{the man admired the agent of the rockers who was/were}). Broken into location of agreement
(High vs.\ Low). Lower surprisal 
corresponds to greater model preference. }
\label{fig:rc_surp}
\end{figure}

\section{Interactions with Syntactic Attachment}

We turn now to the relationship between IC verbs
and syntax. Recall the prediction that object-biased 
IC verbs should interact with RC attachment to license 
more cases of syntactic attachment to the higher noun
compared to lower noun (i.e.\ \textit{chef} in \textit{Anna
scolded the chef of the aristocrats who was/were...}). 

\subsection{Syntactic Stimuli}

We used stimuli from \citet{rohdeetal2011}, which consisted of two experiments: a sentence completion task and a 
self-paced reading task. The sentence completion task consisted of 21 prompts like: 

\ex. \label{2_ic_rc}
    \a. \label{2_ic_rc_a} Carl admires the agent of the rockstars who...
    \b. \label{2_ic_rc_b} Carl works with the agent of the \\rockstars who...

The key manipulation lies with the main verb. In \ref{2_ic_rc_a}, \textit{admires} is an 
object-biased IC verb, and in \ref{2_ic_rc_b}, \textit{works with} is a non-IC verb. 
In the 
present study, we took 14 of these prompts\footnote{7 prompts were excluded because either 
the non-IC or the IC verb was not in the vocabulary of our LSTM LMs. For the remaining prompts, 
we replaced ceo(s) with boss(es), supermodel(s) with superstar(s), and rockstar(s) with rocker(s).}
and generated stimuli balanced for number (i.e.\ we added \textit{Carl admires the agents of the
rockstar who...}), for a total of 112 sentences. 

The self-paced reading time study in \citet{rohdeetal2011} consisted of 20 pairs of sentences, as in:

\ex. \label{rt_ic}
    \a. \label{rt_ic_a} Anna scolded the chef of the aristocrats who was/were routinely 
    letting food go to waste. 
    \b. \label{rt_ic_b} Anna studied with the chef of the aristocrats who was/were routinely 
    letting food go to waste. 
    
As with the completion study, the central manipulation in the self-paced reading study
lies with whether the verb is an 
object-biased IC verb (\textit{scolded}) or not (\textit{studied with}). Rather than give 
completions, though, human participants read sentences where the RC verb (e.g., \textit{was} or 
\textit{were}) either agreed with the higher noun (e.g., \textit{chef}) or the 
lower noun (e.g., \textit{aristocrats}). \citet{rohdeetal2011} reported decreased reading times for agreement with the higher noun when the verb was object-biased compared to 
when the verb was not object-biased. In other words, an object-biased IC verb facilitated 
attachment to the higher noun. In evaluating 
our models on these stimuli, we 
again balanced them by number, so 
that the higher and lower noun were equally frequent
as singular or plural in our test data. This resulted in 192 test sentences generated from 12 pairs.\footnote{We excluded pairs where either of the main verbs was 
not in the vocabulary of our LSTM LMs. There was one noun substitution,
florist(s) with clerk(s). Given that all our LMs were unidirectional, we 
ignored the material after the RC verb. Additionally, for both the completion and self-paced reading stimuli, we substituted male names with 
\textit{the man} and female names with \textit{the woman}.}

\subsection{Measures}

For the sentence completion stimuli, 
we conducted a cloze task. Specifically, given the sentence fragment \textit{the man admires the 
agent of the rockstars who}, we calculated the top 100 most likely next words for 
the LMs. These were then tagged for part-of-speech using \href{https://spacy.io/}{spaCy} and a score was assigned
based on the weighted probability of a continuation using a singular verb (i.e.\ probability 
mass assigned to singular verbs divided by probability mass assigned to all verbs).\footnote{We excluded verbs that were
ambiguous (e.g., \textit{ate}).} Our prediction 
is that object-biased IC main verbs will lead to more continuations agreeing with the 
higher noun (e.g., \textit{agent}).

As detailed in Section \ref{sec:measures}, we calculated information-theoretic surprisal 
and layer-wise similarity. With 
the self-paced reading time stimuli (e.g., \pref{rt_ic}), we calculated surprisal at 
the RC verb and calculated similarity 
between \textit{who} (and \textit{was/were}) and the higher and lower nouns (e.g., 
\textit{chef} and \textit{aristocrats} for \ref{rt_ic}). We predicted that with object-biased IC verbs like \textit{scolded} in \ref{rt_ic}
there would be greater similarity between \textit{who} and \textit{chef} than for 
\textit{who} and \textit{aristocrats} in layers that have an IC distinction (vice versa for non-IC verbs
like \textit{studied}). 

\begin{figure*}[t]
\centering
\includegraphics[width=\textwidth]{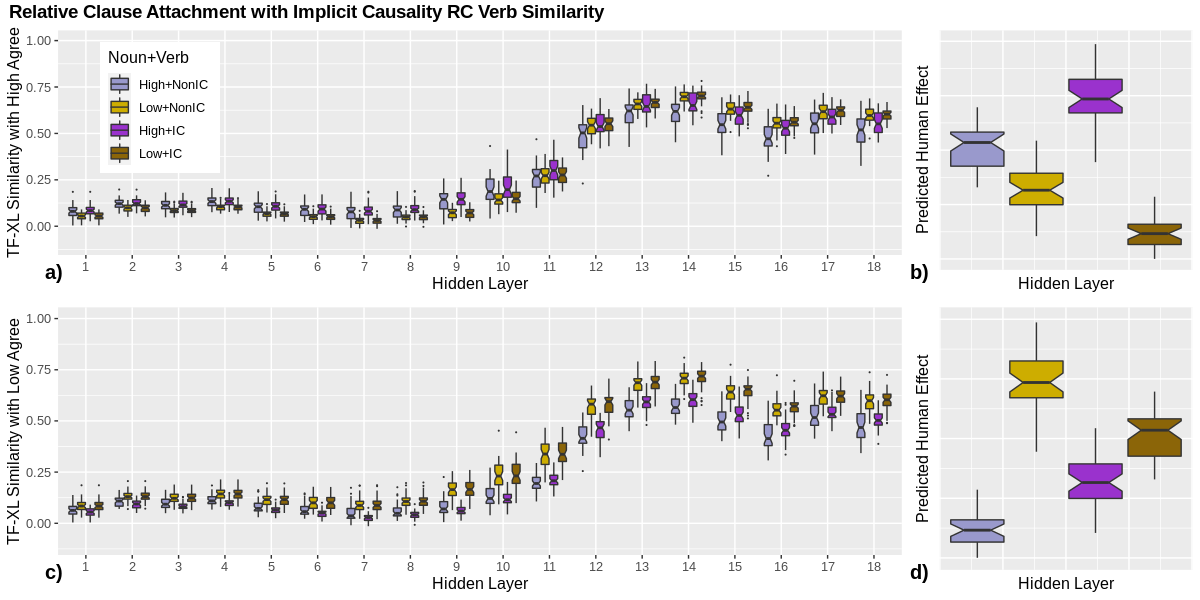}
\caption{Layer-wise representational similarity between 
the RC verb (\textit{was/were}) and the higher/lower noun; stimuli from \citet{rohdeetal2011} (e.g., \textit{the man admired the agent of the rockers who was/were}). Results broken into 
attachment location (higher noun vs.\ lower noun) 
and verb type (object-biased IC verb vs.\ non-IC verb)
are given in \textbf{a)}, for stimuli where the RC verb agrees with the higher 
noun (e.g., \textit{agent of the rockers who was}), and
in \textbf{c)}, for stimuli where the RC verb 
agrees with the lower noun (e.g., \textit{rockers who were}).
The explicit agreement should force a particular attachment location 
to be preferred, with verb IC bias dampening this effect (the predicted 
qualitative human-like pattern is depicted 
in \textbf{b)} and \textbf{d)}). Greater similarity corresponds
to greater relationship between attachment location and \textit{was/were}.}

\label{fig:rc_were_sim}
\end{figure*}

\subsection{Influence of IC on Syntactic Behavior}

To test the influence of IC verbs on model behavior for RC attachment, 
we followed 
the experiments in \citet{rohdeetal2011}. The results are given in Figure \ref{fig:rc_surp}.
We evaluated statistical significance with linear-mixed 
effects models.\footnote{We fit models predicting surprisal at the RC verb 
from a three way interaction of agreement
location, main verb type (object-biased IC or not), and number
and a random intercept for item. For the cloze task, we fit models predicting percent singular 
continuation from an 
interaction between location
of singular agreement (higher or lower noun) and main 
verb type and a random intercept for item.} Post-hoc t-tests
were conducted to assess any effects.
None of the LM architectures showed 
any influence of IC on model behavior, either for the cloze task 
or the self-paced reading stimuli. Rather, 
they all had a strong preference 
for agreeing with the lower noun.\footnote{With regard to categorical preferences (i.e.\ numerically lower surprisal for one attachment location over another for a given stimulus), all the LMs have overwhelming preferences for attachment to the lower noun. The LSTM LMs favored attachment to the higher noun in 0\% of stimuli (across both IC and non-IC stimuli). For TransformerXL, attachment to the higher noun was preferred in 25\% of the stimuli with an object-biased IC verb (i.e.\ where we expect a preference for attachment to the higher noun) and attachment to the higher noun in 27\% of the stimuli 
without an object-biased IC verb (i.e.\ where we do not expect attachment to the higher noun). Finally, for GPT-2 XL, for the stimuli with an object-biased IC verb attachment to the higher noun was preferred in 23\% of the stimuli, and for stimuli without an object-biased IC verbs 9\% of the time.} 

\subsection{Influence of IC on Syntactic Representations} \label{sec:rc_rep}

We examined the representational similarity between 
\textit{who} and the possible attachment points (i.e.\ 
higher or lower noun) and the RC verb (\textit{was/were})
and the possible attachment points. Results for all three 
LM architectures for \textit{who} are given in Figure 
\ref{fig:rc_sim}, and results for the RC verb are given
for TransformerXL in Figure \ref{fig:rc_were_sim}. Statistical
significance was determined via linear-mixed effects 
models.\footnote{Specifically, we fit a model 
predicting the similarity between \textit{who} and 
the possible nouns with a three way interaction of 
main verb type (object-biased IC verb or not), 
noun (higher or lower), and layer with 
a random intercept for item. Additionally, we
fit a model predicting the similarity between \textit{was/were} and 
the possible nouns with a four way interaction of 
main verb type (object-biased IC verb or not), 
noun (higher or lower), agreement location, and layer with a 
random intercept for item.}
Post-hoc t-tests were conducted to assess effects.

The LSTM LMs had no representational effect of IC 
on either \textit{who} or the RC verb, similar to 
the lack of an effect in pronouns. Instead, 
the LSTM LMs representations always had greater similarity 
to the lower noun, in line with the robust preference for attaching 
to the lower noun in behavior (i.e.\ model surprisal). 

For TransformerXL, object-biased IC 
verbs increased the similarity between the 
higher noun and both \textit{who} and 
the RC verb (regardless of agreement). That is, the 
presence of an object-biased IC verb increased the similarity 
of the RC verb and the higher noun both when the RC verb 
agreed in number with the higher noun (e.g., \textit{chef}...\textit{was})
and when the RC verb did not agree in number (e.g., \textit{chef}...\textit{were}). There was no effect 
of IC on the similarity between the lower noun and \textit{who} or
the RC verb.

For
GPT-2, object-biased IC verbs increased
the similarity between the higher noun 
and \textit{who}, but only increased the similarity
between the RC verb and the higher noun 
when they agreed in number (i.e.\ increased similarity 
between \textit{chef} and \textit{was} not \textit{chef} and
\textit{were}). As with TransformerXL, there 
was no analogous effect
on the similarity between the lower noun 
and \textit{who} or 
the RC verb (i.e.\ no change in similarity based on the main verb).

We found TransformerXL had greater 
similarity between the RC verb and
the lower noun in the final layer, regardless 
of verbal agreement (i.e.\ even in cases of ungrammatical attachment, 
TransformerXL preferred local attachment). Similarly, 
GPT-2 XL showed no preference for attachment location 
in the final layers despite
unambiguous agreement with only one of the nouns. Strikingly, both transformer LMs showed greater similarity 
with the agreeing noun (i.e.\ similarity conditioned 
on syntax) in their earlier layers, with the final layers 
obscuring this distinction. 

These results suggest that a preference for local agreement
is robust in both LSTMs and transformer LMs. The transformers
showed representations that encoded the IC contrast, as 
with the referential experiments. However, this knowledge 
did not propagate to the final layers, in line with the absent 
behavioral effects detailed above. Moreover, unambiguous 
syntactic knowledge about RC attachment was discarded 
in the final layers of TransformerXL and GPT-2. These
results suggest that non-linguistic locality preferences 
dominate model representations and behavior. 

\section{Discussion}

The present study examined the extent to which 
discourse structure, determined by implicit causality verbs, 
could be acquired by 
transformer and LSTM language models (cf.\ 
\textit{Sally frightened Mary because she...} and 
\textit{Sally feared Mary because she...}). Specifically, we evaluated,
via comparison to human experiments, whether IC verb biases
could influence reference and syntactic attachment in LMs.
Analyses
were conducted at two levels of granularity: model 
behavior (e.g., probability assigned to possible next words) and model representation (e.g., similarity 
between hidden representations). Given 
the claims in recent literature that implicit causality
arises without extra pragmatic inference on the part 
of human comprehenders, we hypothesized that LMs 
would be able to acquire such contrasts (analogous
to their ability to acquire syntactic agreement). 

We found that LSTM LMs were unable to demonstrate 
knowledge of IC either in influencing reference or syntax. 
However, a 
transformer (TransformerXL) trained on the exact 
same data as the LSTM LMs was able to partially represent
an IC distinction, but model output was only influenced
by IC bias when resolving reference, not syntactic attachment. In evaluating a transformer model trained on 
vastly more data (GPT-2 XL), we found a
more robust, human-like sensitivity to IC bias when 
resolving reference: subject-biased IC verbs 
increased model preference for subject pronouns and object-biased IC verbs
increased model preferences for object pronouns. However, the same mismatch as TransformerXL
between model representation and model behavior 
arose in processing syntactic attachment. 

In contrast to our results, 
\citet{davis-van-schijndel-2020b} showed syntactic predictions for LSTM LMs are influenced by some aspects of
discourse structure. A simple explanation for these conflicting results may 
be that the LMs we examined here are unable to learn the syntactic operation of attachment, and thus no influence of discourse can 
surface. The erasure 
of number agreement in the final layers
of the transformer LMs (see Section \ref{sec:rc_rep}) provides 
compelling evidence towards this conclusion.\footnote{Further cross-linguistic evidence bearing on the inability of LSTM LMs, specifically, to learn relative clause attachment is given in \citet{davis-van-schijndel-2020-recurrent}.} 

From a theoretical perspective, the present study
provides additional support for the centering of 
implicit causality within the linguistic signal proper. 
That is, IC bias is learnable, to some degree, 
without pragmatic inference as hypothesized in Section \ref{sec:intro} \cite[see also][]{hartshorne2014}. 
The mismatches 
in syntactic representations and behavior suggest, however, that models
ignore the abstract categories that are learned, 
contrary to human findings \cite[cf.][]{rohdeetal2011}. 

We believe a solution may lie in changing model 
training objectives
(i.e.\ what linguistic unit should be predicted). 
Psycholinguistic studies focusing on the interaction 
of discourse and syntax have suggested that 
coherence relations may be the unit of linguistic 
prediction, in contrast to the next-word prediction 
used in most language modeling work \cite[see][]{rohdeetal2011}. We leave to 
future work an 
investigation of this suggestion as well as 
teasing apart the exact role that training data and 
model architecture play in the interaction 
between types of linguistic representation.

\section*{Acknowledgments}
Thank you to members of the C.Psyd lab at Cornell, who gave feedback on an earlier form of this work. We would also like to thank the three anonymous reviewers for their comments and suggestions.

\bibliography{conll2020}

\begin{thebibliography}{46}
\expandafter\ifx\csname natexlab\endcsname\relax\def\natexlab#1{#1}\fi

\bibitem[{Bates et~al.(2015)Bates, M{\"a}chler, Bolker, and
  Walker}]{bates2015lme4}
Douglas Bates, Martin M{\"a}chler, Ben Bolker, and Steve Walker. 2015.
\newblock \href {https://doi.org/10.18637/jss.v067.i01} {Fitting linear
  mixed-effects models using {lme4}}.
\newblock \emph{Journal of Statistical Software}, 67(1):1--48.

\bibitem[{Bernardy and Lappin(2017)}]{bernardyetal2017}
Jean-Philippe Bernardy and Shalom Lappin. 2017.
\newblock Using deep neural networks to learn syntactic agreement.
\newblock \emph{Linguistic Issues in Language Technology}, 15.

\bibitem[{Cheng and Erk(2020)}]{cheng-erk-2020-attending}
Pengxiang Cheng and Katrin Erk. 2020.
\newblock \href {https://arxiv.org/abs/1911.04361} {{Attending to Entities for
  Better Text Understanding}}.
\newblock In \emph{Proceedings of the {AAAI} Conference on Artificial
  Intelligence}, volume~34.

\bibitem[{Chrupa{\l}a and Alishahi(2019)}]{chrupala-alishahi-2019-correlating}
Grzegorz Chrupa{\l}a and Afra Alishahi. 2019.
\newblock \href {https://doi.org/10.18653/v1/P19-1283} {Correlating neural and
  symbolic representations of language}.
\newblock In \emph{Proceedings of the 57th Annual Meeting of the Association
  for Computational Linguistics}, pages 2952--2962, Florence, Italy.
  Association for Computational Linguistics.

\bibitem[{Dai et~al.(2019)Dai, Yang, Yang, Carbonell, Le, and
  Salakhutdinov}]{dai-etal-2019-transformer}
Zihang Dai, Zhilin Yang, Yiming Yang, Jaime Carbonell, Quoc Le, and Ruslan
  Salakhutdinov. 2019.
\newblock \href {https://doi.org/10.18653/v1/P19-1285} {{Transformer-{XL}:
  Attentive Language Models beyond a Fixed-Length Context}}.
\newblock In \emph{Proceedings of the 57th Annual Meeting of the Association
  for Computational Linguistics}, pages 2978--2988, Florence, Italy.
  Association for Computational Linguistics.

\bibitem[{Davis and van
  Schijndel(2020{\natexlab{a}})}]{davis-van-schijndel-2020b}
Forrest Davis and Marten van Schijndel. 2020{\natexlab{a}}.
\newblock \href {https://doi.org/10.31234/osf.io/8r65d} {{Interaction with
  Context During Recurrent Neural Network Sentence Processing}}.
\newblock In \emph{Proceedings of the 42nd Annual Meeting of the Cognitive
  Science Society}.

\bibitem[{Davis and van
  Schijndel(2020{\natexlab{b}})}]{davis-van-schijndel-2020-recurrent}
Forrest Davis and Marten van Schijndel. 2020{\natexlab{b}}.
\newblock \href {https://www.aclweb.org/anthology/2020.acl-main.179}
  {{Recurrent Neural Network Language Models Always Learn {E}nglish-Like
  Relative Clause Attachment}}.
\newblock In \emph{Proceedings of the 58th Annual Meeting of the Association
  for Computational Linguistics}, pages 1979--1990, Online. Association for
  Computational Linguistics.

\bibitem[{Devlin et~al.(2019)Devlin, Chang, Lee, and Toutanova}]{devlinetal19}
Jacob Devlin, Ming-Wei Chang, Kenton Lee, and Kristina Toutanova. 2019.
\newblock \href {https://arxiv.org/pdf/1810.04805.pdf} {{BERT}: {Pre-training
  of Deep Bidirectional Transformers for Language Understanding}}.
\newblock In \emph{Proceedings of the 2019 Annual Conference of the North
  American Chapter of the Association for Computational Linguistics}.
  Association for Computational Linguistics.

\bibitem[{Enguehard et~al.(2017)Enguehard, Goldberg, and Linzen}]{Enguehard17}
{\'E}mile Enguehard, Yoav Goldberg, and Tal Linzen. 2017.
\newblock \href {https://doi.org/10.18653/v1/K17-1003} {{Exploring the
  Syntactic Abilities of {RNNs} with Multi-task Learning}}.
\newblock In \emph{Proceedings of the 21st Conference on Computational Natural
  Language Learning (CoNLL 2017)}, pages 3--14. Association for Computational
  Linguistics.

\bibitem[{Ferstl et~al.(2011)Ferstl, Garnham, and
  Manouilidou}]{ferstl2011implicit}
Evelyn~C Ferstl, Alan Garnham, and Christina Manouilidou. 2011.
\newblock \href {https://link.springer.com/article/10.3758/s13428-010-0023-2}
  {Implicit causality bias in {E}nglish: A corpus of 300 verbs}.
\newblock \emph{Behavior Research Methods}, 43(1):124--135.

\bibitem[{{Frank} et~al.(2015){Frank}, {Otten}, {Galli}, and
  {Vigliocco}}]{franketal15}
Stefan~L. {Frank}, Leun~J. {Otten}, Giulia {Galli}, and Gabriella {Vigliocco}.
  2015.
\newblock \href {https://doi.org/10.1016/j.bandl.2014.10.006} {{The ERP
  response to the amount of information conveyed by words in sentences}}.
\newblock \emph{Brain \& Language}, 140:1--11.

\bibitem[{Futrell et~al.(2018)Futrell, Wilcox, Morita, and
  Levy}]{futrelletal2018}
Richard Futrell, Ethan Wilcox, Takashi Morita, and Roger Levy. 2018.
\newblock \href {https://arxiv.org/abs/1809.01329} {{RNN}s as psycholinguistic
  subjects: {S}yntactic state and grammatical dependency}.
\newblock \emph{arXiv preprint arXiv:1809.01329}.

\bibitem[{Garvey and Caramazza(1974)}]{garvey1974implicit}
Catherine Garvey and Alfonso Caramazza. 1974.
\newblock Implicit causality in verbs.
\newblock \emph{Linguistic inquiry}, 5(3):459--464.

\bibitem[{Giulianelli et~al.(2018)Giulianelli, Harding, Mohnert, Hupkes, and
  Zuidema}]{giulianelli-etal-2018-hood}
Mario Giulianelli, Jack Harding, Florian Mohnert, Dieuwke Hupkes, and Willem
  Zuidema. 2018.
\newblock \href {https://doi.org/10.18653/v1/W18-5426} {Under the hood: Using
  diagnostic classifiers to investigate and improve how language models track
  agreement information}.
\newblock In \emph{Proceedings of the 2018 {EMNLP} Workshop {B}lackbox{NLP}:
  Analyzing and Interpreting Neural Networks for {NLP}}, pages 240--248,
  Brussels, Belgium. Association for Computational Linguistics.

\bibitem[{Gulordava et~al.(2018)Gulordava, Bojanowski, Grave, Linzen, and
  Baroni}]{gulordavaetal18}
Kristina Gulordava, Piotr Bojanowski, Edouard Grave, Tal Linzen, and Marco
  Baroni. 2018.
\newblock \href {https://www.aclweb.org/anthology/N18-1108} {Colorless green
  recurrent networks dream hierarchically}.
\newblock In \emph{Proceedings of the 2018 Annual Conference of the North
  American Chapter of the Association for Computational Linguistics}.
  Association for Computational Linguistics.

\bibitem[{Hale(2001)}]{hale2001probabilistic}
John Hale. 2001.
\newblock \href {https://www.aclweb.org/anthology/N01-1021/} {A probabilistic
  {Earley} parser as a psycholinguistic model}.
\newblock In \emph{Proceedings of the second meeting of the North American
  Chapter of the Association for Computational Linguistics on Language
  technologies}, pages 1--8. Association for Computational Linguistics.

\bibitem[{Hartshorne(2014)}]{hartshorne2014}
Joshua~K Hartshorne. 2014.
\newblock \href {https://doi.org/10.1080/01690965.2013.796396} {What is
  implicit causality?}
\newblock \emph{Language, Cognition and Neuroscience}, 29(7):804--824.

\bibitem[{Hartshorne and Snedeker(2013)}]{hartshorne2013verb}
Joshua~K Hartshorne and Jesse Snedeker. 2013.
\newblock \href {https://doi.org/10.1080/01690965.2012.689305} {Verb argument
  structure predicts implicit causality: The advantages of finer-grained
  semantics}.
\newblock \emph{Language and Cognitive Processes}, 28(10):1474--1508.

\bibitem[{{Hochreiter} and {Schmidhuber}(1997)}]{hochreiterschmidhuber97}
Sepp {Hochreiter} and J\"{u}rgen {Schmidhuber}. 1997.
\newblock \href
  {https://www.mitpressjournals.org/doi/10.1162/neco.1997.9.8.1735} {Long
  short-term memory}.
\newblock \emph{Neural Computation}, 9(8):1735--1780.

\bibitem[{Hu et~al.(2020)Hu, Gauthier, Qian, Wilcox, and
  Levy}]{huetal2020-systematic}
Jennifer Hu, Jon Gauthier, Peng Qian, Ethan Wilcox, and Roger Levy. 2020.
\newblock \href {https://www.aclweb.org/anthology/2020.acl-main.158} {{A
  Systematic Assessment of Syntactic Generalization in Neural Language
  Models}}.
\newblock In \emph{Proceedings of the 58th Annual Meeting of the Association
  for Computational Linguistics}, pages 1725--1744, Online. Association for
  Computational Linguistics.

\bibitem[{Jeretic et~al.(2020)Jeretic, Warstadt, Bhooshan, and
  Williams}]{jeretic-etal-2020-natural}
Paloma Jeretic, Alex Warstadt, Suvrat Bhooshan, and Adina Williams. 2020.
\newblock \href {https://doi.org/10.18653/v1/2020.acl-main.768} {Are natural
  language inference models {IMPPRESsive}? {L}earning {IMPlicature} and
  {PRESupposition}}.
\newblock In \emph{Proceedings of the 58th Annual Meeting of the Association
  for Computational Linguistics}, pages 8690--8705, Online. Association for
  Computational Linguistics.

\bibitem[{Jumelet et~al.(2019)Jumelet, Zuidema, and
  Hupkes}]{jumelet-etal-2019-analysing}
Jaap Jumelet, Willem Zuidema, and Dieuwke Hupkes. 2019.
\newblock \href {https://doi.org/10.18653/v1/K19-1001} {{Analysing Neural
  Language Models: Contextual Decomposition Reveals Default Reasoning in Number
  and Gender Assignment}}.
\newblock In \emph{Proceedings of the 23rd Conference on Computational Natural
  Language Learning}, pages 1--11, Hong Kong, China. Association for
  Computational Linguistics.

\bibitem[{Kehler et~al.(2008)Kehler, Kertz, Rohde, and
  Elman}]{kehler2008coherence}
Andrew Kehler, Laura Kertz, Hannah Rohde, and Jeffrey~L Elman. 2008.
\newblock \href {https://doi.org/10.1093/jos/ffm018} {Coherence and coreference
  revisited}.
\newblock \emph{Journal of semantics}, 25(1):1--44.

\bibitem[{Kodner and Gupta(2020)}]{kodner-gupta-2020-overestimation}
Jordan Kodner and Nitish Gupta. 2020.
\newblock \href {https://www.aclweb.org/anthology/2020.acl-main.160}
  {{Overestimation of Syntactic Representation in Neural Language Models}}.
\newblock In \emph{Proceedings of the 58th Annual Meeting of the Association
  for Computational Linguistics}, pages 1757--1762, Online. Association for
  Computational Linguistics.

\bibitem[{Kriegeskorte et~al.(2008)Kriegeskorte, Mur, and
  Bandettini}]{kriegeskorte2008representational}
Nikolaus Kriegeskorte, Marieke Mur, and Peter~A Bandettini. 2008.
\newblock \href {https://doi.org/10.3389/neuro.06.004.2008} {Representational
  similarity analysis-connecting the branches of systems neuroscience}.
\newblock \emph{Frontiers in Systems Neuroscience}, 2:4.

\bibitem[{Kuznetsova et~al.(2017)Kuznetsova, Brockhoff, and
  Christensen}]{kuznetsova2017lmertest}
Alexandra Kuznetsova, Per~B. Brockhoff, and Rune H.~B. Christensen. 2017.
\newblock \href {https://doi.org/10.18637/jss.v082.i13} {{lmerTest} package:
  Tests in linear mixed effects models}.
\newblock \emph{Journal of Statistical Software}, 82(13):1--26.

\bibitem[{Levy(2008)}]{levy2008expectation}
Roger Levy. 2008.
\newblock \href {https://doi.org/10.1016/j.cognition.2007.05.006}
  {Expectation-based syntactic comprehension}.
\newblock \emph{Cognition}, 106(3):1126--1177.

\bibitem[{Linzen et~al.(2016)Linzen, Dupoux, and Goldberg}]{linzenetal16}
Tal Linzen, Emmanuel Dupoux, and Yoav Goldberg. 2016.
\newblock \href {https://www.aclweb.org/anthology/Q16-1037} {Assessing the
  ability of {LSTMs} to learn syntax-sensitive dependencies}.
\newblock \emph{Transactions of the Association for Computational Linguistics},
  4:521--535.

\bibitem[{Merity et~al.(2016)Merity, Xiong, Bradbury, and
  Socher}]{merityetal16}
Stephen Merity, Caiming Xiong, James Bradbury, and Richard Socher. 2016.
\newblock Wikitext-103.
\newblock Technical report, Salesforce.

\bibitem[{Morcos et~al.(2018)Morcos, Raghu, and Bengio}]{morcos2018insights}
Ari Morcos, Maithra Raghu, and Samy Bengio. 2018.
\newblock \href
  {http://papers.nips.cc/paper/7815-insights-on-representational-similarity-in-neural-networks-with-canonical-correlation.pdf}
  {Insights on representational similarity in neural networks with canonical
  correlation}.
\newblock In S.~Bengio, H.~Wallach, H.~Larochelle, K.~Grauman, N.~Cesa-Bianchi,
  and R.~Garnett, editors, \emph{Advances in Neural Information Processing
  Systems 31}, pages 5727--5736. Curran Associates, Inc.

\bibitem[{Mueller et~al.(2020)Mueller, Nicolai, Petrou-Zeniou, Talmina, and
  Linzen}]{mueller-etal-2020-cross}
Aaron Mueller, Garrett Nicolai, Panayiota Petrou-Zeniou, Natalia Talmina, and
  Tal Linzen. 2020.
\newblock \href {https://www.aclweb.org/anthology/2020.acl-main.490}
  {{Cross-Linguistic Syntactic Evaluation of Word Prediction Models}}.
\newblock In \emph{Proceedings of the 58th Annual Meeting of the Association
  for Computational Linguistics}, pages 5523--5539, Online. Association for
  Computational Linguistics.

\bibitem[{Peters et~al.(2018)Peters, Neumann, Iyyer, Gardner, Clark, Lee, and
  Zettlemoyer}]{petersetal18}
Matthew~E. Peters, Mark Neumann, Mohit Iyyer, Matt Gardner, Christopher Clark,
  Kenton Lee, and Luke Zettlemoyer. 2018.
\newblock \href {https://aclweb.org/anthology/N18-1202} {Deep contextualized
  word representations}.
\newblock In \emph{Proceedings of the 2018 Annual Conference of the North
  American Chapter of the Association for Computational Linguistics}.
  Association for Computational Linguistics.

\bibitem[{Prasad et~al.(2019)Prasad, {van Schijndel}, and
  Linzen}]{prasadetal2019}
Grusha Prasad, Marten {van Schijndel}, and Tal Linzen. 2019.
\newblock \href {https://www.aclweb.org/anthology/K19-1007/} {{Using Priming to
  Uncover the Organization of Syntactic Representations in Neural Language
  Models}}.
\newblock In \emph{Proceedings of the 23rd Conference on Computational Natural
  Language Learning}.

\bibitem[{Radford et~al.(2018)Radford, Narasimhan, Salimans, and
  Sutskever}]{radfordetal18}
Alec Radford, Karthik Narasimhan, Tim Salimans, and Ilya Sutskever. 2018.
\newblock \href
  {https://s3-us-west-2.amazonaws.com/openai-assets/research-covers/language-unsupervised/language_understanding_paper.pdf}
  {{Improving Language Understanding by Generative Pre-Training}}.
\newblock Technical report, OpenAI.

\bibitem[{Radford et~al.(2019)Radford, Wu, Child, Luan, Amodei, and
  Sutskever}]{radfordetal19}
Alec Radford, Jeffrey Wu, Rewon Child, David Luan, Dario Amodei, and Ilya
  Sutskever. 2019.
\newblock \href
  {https://d4mucfpksywv.cloudfront.net/better-language-models/language_models_are_unsupervised_multitask_learners.pdf}
  {{Language Models are Unsupervised Multitask Learners}}.
\newblock Technical report, OpenAI.

\bibitem[{Rohde et~al.(2011)Rohde, Levy, and Kehler}]{rohdeetal2011}
Hannah Rohde, Roger Levy, and Andrew Kehler. 2011.
\newblock \href {https://doi.org/10.1016/j.cognition.2010.10.016} {Anticipating
  explanations in relative clause processing}.
\newblock \emph{Cognition}, 118(3):339--358.

\bibitem[{Schuster et~al.(2020)Schuster, Chen, and
  Degen}]{schuster-etal-2020-harnessing}
Sebastian Schuster, Yuxing Chen, and Judith Degen. 2020.
\newblock \href {https://www.aclweb.org/anthology/2020.acl-main.479}
  {Harnessing the linguistic signal to predict scalar inferences}.
\newblock In \emph{Proceedings of the 58th Annual Meeting of the Association
  for Computational Linguistics}, pages 5387--5403, Online. Association for
  Computational Linguistics.

\bibitem[{{Shannon}(1948)}]{shannon48}
Claude {Shannon}. 1948.
\newblock \href {https://doi.org/10.1002/j.1538-7305.1948.tb01338.x} {A
  mathematical theory of communication}.
\newblock \emph{Bell System Technical Journal}, 27:379--423, 623--656.

\bibitem[{Smith and Levy(2013)}]{smith2013effect}
Nathaniel~J Smith and Roger Levy. 2013.
\newblock \href {https://doi.org/10.1016/j.cognition.2013.02.013} {The effect
  of word predictability on reading time is logarithmic}.
\newblock \emph{Cognition}, 128(3):302--319.

\bibitem[{Sorodoc et~al.(2020)Sorodoc, Gulordava, and
  Boleda}]{sorodoc-etal-2020-probing}
Ionut-Teodor Sorodoc, Kristina Gulordava, and Gemma Boleda. 2020.
\newblock \href {https://www.aclweb.org/anthology/2020.acl-main.384} {{Probing
  for Referential Information in Language Models}}.
\newblock In \emph{Proceedings of the 58th Annual Meeting of the Association
  for Computational Linguistics}, pages 4177--4189, Online. Association for
  Computational Linguistics.

\bibitem[{Trask et~al.(2018)Trask, Hill, Reed, Rae, Dyer, and
  Blunsom}]{trasketal2018}
Andrew Trask, Felix Hill, Scott~E Reed, Jack Rae, Chris Dyer, and Phil Blunsom.
  2018.
\newblock \href {https://arxiv.org/abs/1808.00508} {Neural arithmetic logic
  units}.
\newblock In \emph{Advances in Neural Information Processing Systems}, pages
  8035--8044.

\bibitem[{{van Schijndel} et~al.(2019){van Schijndel}, Mueller, and
  Linzen}]{vanSchijndeletal2019}
Marten {van Schijndel}, Aaron Mueller, and Tal Linzen. 2019.
\newblock \href {https://www.aclweb.org/anthology/D19-1592/} {Quantity doesn't
  buy quality syntax with neural language models}.
\newblock In \emph{Proceedings of the 2019 Conference on Empirical Methods in
  Natural Language Processing}. Association for Computational Linguistics.

\bibitem[{Wilcox et~al.(2018)Wilcox, Levy, and Futrell}]{wilcox2019block}
Ethan Wilcox, Roger Levy, and Richard Futrell. 2018.
\newblock \href {https://arxiv.org/abs/1905.10431} {{What Syntactic Structures
  block Dependencies in RNN Language Models?}}
\newblock In \emph{Proceedings of the 41st Annual Meeting of the Cognitive
  Science Society}.

\bibitem[{Wilcox et~al.(2019)Wilcox, Levy, and Futrell}]{wilcox2019supression}
Ethan Wilcox, Roger Levy, and Richard Futrell. 2019.
\newblock \href {https://www.aclweb.org/anthology/W19-4819/} {{Hierarchical
  Representation in Neural Language Models: Suppression and Recovery of
  Expectations}}.
\newblock In \emph{Proceedings of the 2019 ACL Workshop BlackboxNLP: Analyzing
  and Interpreting Neural Networks for NLP}.

\bibitem[{Williams(2020)}]{williams_IC_2020}
Elyce~Dominique Williams. 2020.
\newblock \href {https://doi.org/10.17615/0x4r-rq30} {Language {Experience}
  {Predicts} {Pronoun} {Comprehension} in {Implicit} {Causality} {Sentences}}.
\newblock Master's thesis, University of North Carolina at Chapel Hill.

\bibitem[{Wolf et~al.(2019)Wolf, Debut, Sanh, Chaumond, Delangue, Moi, Cistac,
  Rault, Louf, Funtowicz, and Brew}]{Wolf2019HuggingFacesTS}
Thomas Wolf, Lysandre Debut, Victor Sanh, Julien Chaumond, Clement Delangue,
  Anthony Moi, Pierric Cistac, Tim Rault, R'emi Louf, Morgan Funtowicz, and
  Jamie Brew. 2019.
\newblock \href {https://arxiv.org/abs/1910.03771} {{HuggingFace's
  Transformers: State-of-the-art Natural Language Processing}}.
\newblock \emph{ArXiv}, abs/1910.03771.

\end{thebibliography}
\bibliographystyle{acl_natbib}

\appendix
\section{Stereotypically gendered nouns used in referential experiments}\label{pairs}

    \begin{center}

    \begin{tabular}{c|c}
         \textbf{male} & \textbf{female}\\
         \hline
        man & woman \\
        boy &girl \\
        father& mother\\
        uncle &aunt\\
        husband &wife\\
        actor &actress\\
        prince &princess\\
        waiter &waitress\\
        lord &lady\\
        king &queen\\
        son &daughter\\
        nephew &niece\\
        brother &sister\\
        grandfather& grandmother\\
    \end{tabular}
            
    \end{center}

\end{document}